\documentclass[10pt,twocolumn,letterpaper]{article}

\usepackage[pagenumbers]{cvpr} %

\usepackage{algorithm}
\usepackage{algorithmic}
\usepackage{url}            %
\usepackage{booktabs}       %
\usepackage{amsfonts}       %
\usepackage{amssymb}     
\usepackage{amsmath}     
\usepackage{nicefrac}       %
\usepackage{microtype}      %
\usepackage{xcolor}         %
\usepackage{bm}     
\usepackage{xfrac} 
\usepackage{array}
\usepackage{makecell}
\usepackage{multirow}
\usepackage{enumitem}
\usepackage{color}
\usepackage{colortbl}

\definecolor{lightblue}{RGB}{122,46,246}
\definecolor{lightgreen}{RGB}{0,128,0}
\definecolor{lightred}{RGB}{226,3,9}
\definecolor{lightorange}{RGB}{225,165,0}
\definecolor{lightours}{RGB}{255,246,247}
\definecolor{FLOPs}{RGB}{248,203,173}

\newcommand{\PreserveBackslash}[1]{\let\temp=\\#1\let\\=\temp}
\newcolumntype{C}[1]{>{\PreserveBackslash\centering}p{#1}}
\newcolumntype{R}[1]{>{\PreserveBackslash\raggedleft}p{#1}}
\newcolumntype{L}[1]{>{\PreserveBackslash\raggedright}p{#1}}

\definecolor{cvprblue}{rgb}{0.21,0.49,0.74}
\usepackage[pagebackref,breaklinks,colorlinks,allcolors=cvprblue]{hyperref}

\def\paperID{*****} %
\def\confName{CVPR}
\def\confYear{2025}

\title{Effective Diffusion Transformer Architecture for Image Super-Resolution}

\author{
    Kun Cheng\textsuperscript{1}\footnotemark[1], Lei Yu\textsuperscript{2}\footnotemark[1], Zhijun Tu\textsuperscript{2}, Xiao He\textsuperscript{1}, Liyu Chen\textsuperscript{2}, Yong Guo\textsuperscript{3}, \\ Mingrui Zhu\textsuperscript{1}, Nannan Wang\textsuperscript{1}\footnotemark[2], Xinbo Gao\textsuperscript{4}, Jie Hu\textsuperscript{2} \\
    \textsuperscript{1}State Key Laboratory of Integrated Services Networks, Xidian University \\
    \textsuperscript{2}Huawei Noah's Ark Lab \\
    \textsuperscript{3}Consumer Business Group, Huawei \\
    \textsuperscript{4}Chongqing Key Laboratory of Image Cognition, Chongqing University of Posts and\\ Telecommunications\\
    \tt\small kunncheng@stu.xidian.edu.cn, yulei96@huawei.com, nnwang@xidian.edu.cn
}

\begin{document}
\maketitle

\renewcommand{\thefootnote}{\fnsymbol{footnote}}
\footnotetext[1]{Both authors contributed equally to this research.}
\footnotetext[2]{Corresponding author.}
\renewcommand{\thefootnote}{\arabic{footnote}}

\begin{abstract}
    Recent advances indicate that diffusion models hold great promise in image super-resolution. While the latest methods are primarily based on latent diffusion models with convolutional neural networks, there are few attempts to explore transformers, which have demonstrated remarkable performance in image generation. In this work, we design an effective diffusion transformer for image super-resolution (DiT-SR) that achieves the visual quality of prior-based methods, but through a training-from-scratch manner. In practice, DiT-SR leverages an overall U-shaped architecture, and adopts a uniform isotropic design for all the transformer blocks across different stages. The former facilitates multi-scale hierarchical feature extraction, while the latter reallocates the computational resources to critical layers to further enhance performance. Moreover, we thoroughly analyze the limitation of the widely used AdaLN, and present a frequency-adaptive time-step conditioning module, enhancing the model's capacity to process distinct frequency information at different time steps. Extensive experiments demonstrate that DiT-SR outperforms the existing training-from-scratch diffusion-based SR methods significantly, and even beats some of the prior-based methods on pretrained Stable Diffusion, proving the superiority of diffusion transformer in image super-resolution.
\end{abstract}

\begin{figure}[t]
    \includegraphics[width=1\linewidth]{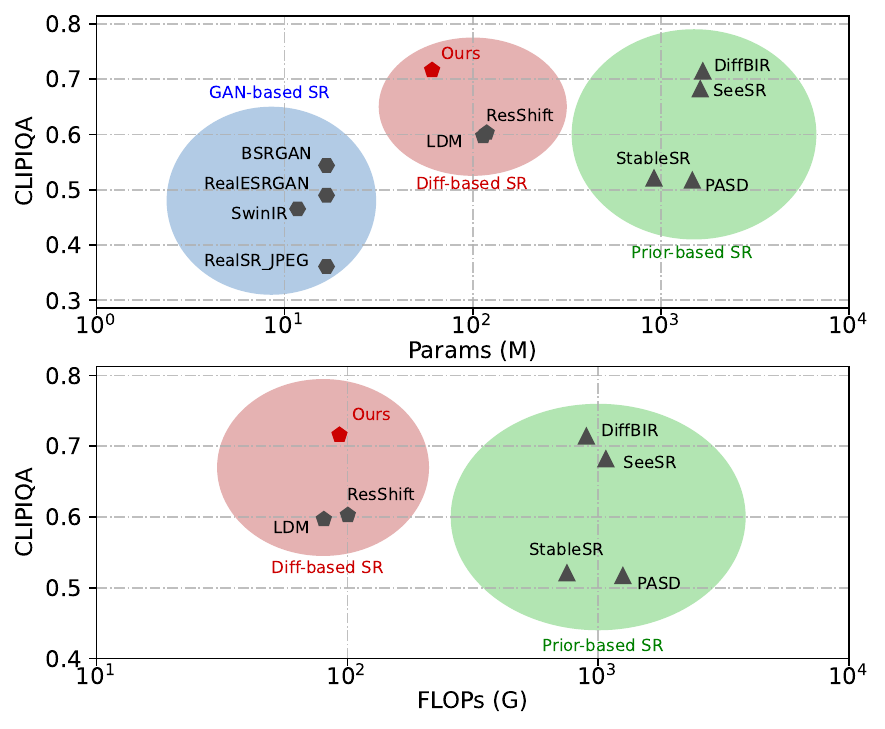}
    \caption{Comparisons between the proposed method and the latest SR methods on RealSR dataset. Top: CLIPIQA vs. Parameters. Bottom: CLIPIQA vs. FLOPs. Specifically, ``Diff-based SR'' refers to diffusion-based image super-resolution methods trained from scratch.}
    \label{fig:teaser}
\end{figure}

\section{Introduction}
Image super-resolution~(SR) aims to reconstruct a high-resolution~(HR) image from a low-resolution~(LR) input. 
Recently, diffusion models (DMs)~\cite{ho2020denoising, dhariwal2021diffusion, rombach2022high} have demonstrated superior performance in image generation. Notable works~\cite{wang2023exploiting,lin2024diffbir,yu2024scaling,wu2024seesr} have applied DMs to image super-resolution, achieving exceptional performance, particularly on complex natural scenes. 
Specifically, diffusion-based SR methods typically fall into two categories: the first group~\cite{saharia2022image,rombach2022high,shang2024resdiff,yue2024resshift} involves injecting LR images directly into the diffusion model and training it from scratch, while the second group~\cite{wang2023exploiting,lin2024diffbir,yang2023pasd,yu2024scaling,wu2024seesr}, exploits the generative prior from pre-trained diffusion models, such as Stable Diffusion~(SD)~\cite{rombach2022high,podell2023sdxl}, to enhance image super-resolution. Methods trained from scratch offer significant flexibility and ease of retraining following architectural modifications, making them ideal for lightweight applications. 
However, as shown in Fig.~\ref{fig:teaser}, these methods typically struggle to match the upper bound performance of prior-based methods, which benefit from the rich generative prior gained through extensive training on vast datasets over thousands of GPU days. 
A natural question is can we develop a diffusion architecture trained from scratch while rivaling the performance of prior-based methods, balancing both performance and flexibility?

The advent of the Diffusion Transformer (DiT)~\cite{peebles2023scalable} has made this idea feasible. This isotropic, full-transformer architecture, which maintains constant resolution and channel dimensions, shows remarkable performance and scalability, establishing a new paradigm in diffusion architecture design~\cite{videoworldsimulators2024,esser2024scaling, li2024hunyuan,gao2024lumina,hatamizadeh2023diffit}. 
In contrast, early diffusion works~\cite{ho2020denoising, dhariwal2021diffusion, rombach2022high} typically employed U-shaped denoiser architecture, which also remains popular in low-level tasks~\cite{wang2022uformer, zamir2021multi} due to its hierarchical feature extraction capability and inductive bias conducive to denoising~\cite{williams2024unified}. 

In this paper, we propose a diffusion transformer model for image super-resolution, namely DiT-SR. Instead of applying the standard diffusion transformer architecture directly, DiT-SR is a U-shape encoder-decoder network, but with isotropic designs for all the transformer blocks at different stages. Specifically, DiT-SR adopts the U-shaped global structure with incrementally wider channel dimensions at deeper layers, which helps recover more image details at multi-scale resolutions. Besides, inspired by the observations that (1) The transformer architecture with the same depth and channels could process tokens of different lengths well, \textit{e.g.}, DiT-XL/2, DiT-XL/4 and DiT-XL/8. (2) High-resolution DiTs (\textit{e.g.}, DiT/2) benefit more from scaling up than low-resolution DiTs (\textit{e.g.}, DiT/8), thus we introduce the isotropic designs of DiT into the multi-scale framework. DiT-SR mandates the same channel number for all transformer modules in different stages, and sets the channel number bigger than the original setting of high resolution in U-Net, but much smaller than the low resolution. By allocating computational resource to critical layers, DiT-SR can greatly boost the capacity of the transformer architecture in multi-scale paradigms with the given computation budget.

\begin{figure}
    \centering
    \setlength{\abovecaptionskip}{2mm}
    \includegraphics[width=1\linewidth]{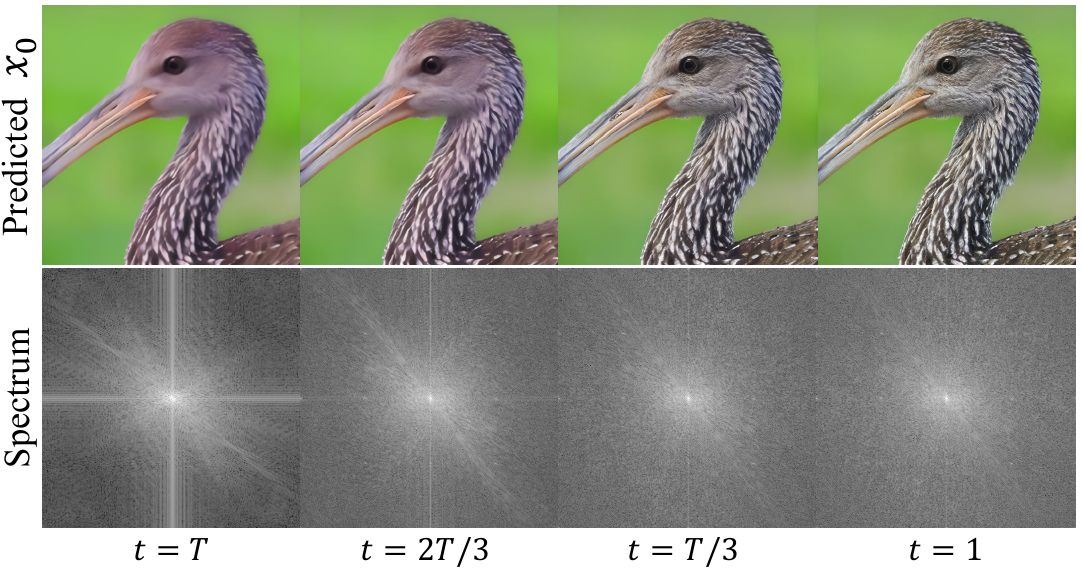}
    \caption{Analysis of images generated at different stages with a diffusion-based super-resolution model~\cite{yue2024resshift}. The first row shows the predicted clean images at various steps, while the second row displays the Fourier spectrums of each predicted clean image. The diffusion model initially generates low-frequency components (center part of spectrums) and subsequently generates high-frequency components (peripheral part of spectrums). }
    \label{fig:sr_freq}
\end{figure}

Furthermore, we observe that DiT-based denoisers encounter a common issue~\cite{hatamizadeh2023diffit} related to the inefficient mechanism of time-step conditioning.  As illustrated in Fig.~\ref{fig:sr_freq}, the diffusion-based SR model attends to different frequency components at distinct denoising phases. Consequently, there should be a direct correlation between the time step and frequency. However, the widespreadly used Adaptive Layer Normalization (AdaLN), which modulates features solely in a channel-wise manner, does not effectively capture the unique temporal dynamics of the denoising process.
To overcome this limitation, we propose an Adaptive Frequency Modulation~(AdaFM) module, conditioning on the frequency domain. This highly efficient module, replacing AdaLN after each normalization layer, requires significantly fewer parameters while boosting performance. The time step adaptively reweights different frequency components, making it especially suitable for image super-resolution, which necessitates a strong emphasis on high-frequency details. 

We summarize the primary contributions as follows:
\begin{itemize}[topsep=0pt,parsep=0pt,leftmargin=18pt]
    \item We propose DiT-SR, a diffusion transformer specifically designed for image super-resolution, the first work that seamlessly combines the advantages of U-shaped and isotropic designs.
    \item We introduce an efficient yet effective frequency-wise time step conditioning module AdaFM, augmenting the diffusion model's ability to emphasize specific frequency information at varying time steps.
    \item Extensive experiments demonstrate that the proposed diffusion architecture outperforms existing training-from-scratch  SR methods dramatically, and even surpasses some of prior-based SR methods with about only $ 5\% $ of the parameters. 
\end{itemize}

\begin{figure*}
    \centering
    \setlength{\abovecaptionskip}{2mm}
    \includegraphics[width=1\linewidth]{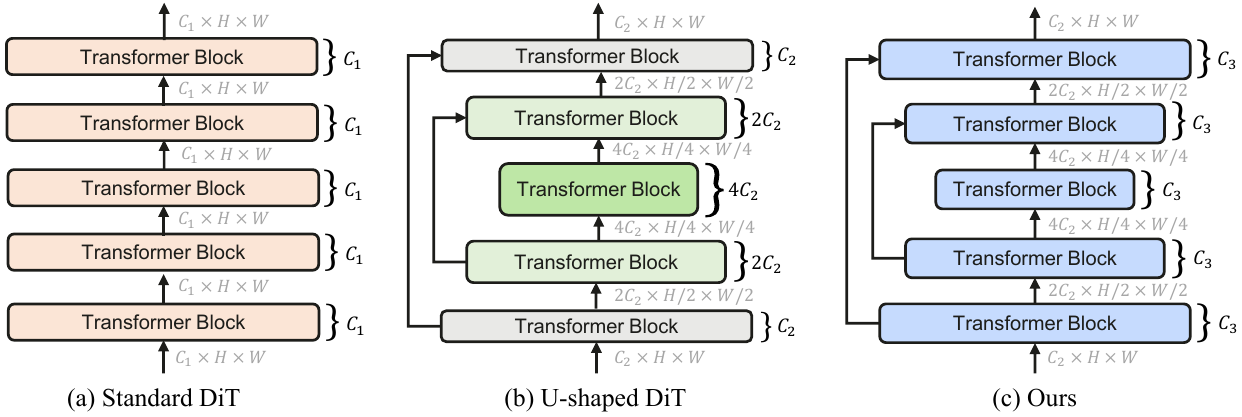}
    \caption{The comparison from the standard DiT to the proposed DiT-SR. (a): The standard DiT. (b):U-shaped DiT, incorporating downsampling and upsampling to standard DiT and increasing the channel dimension in deep layers. (c): The proposed DiT-SR. This architecture employs a U-shaped global structure, yet maintains the same channel dimension for all transformer blocks in different stages, allocating computational resource to high-resolution layers~($4C_2 > C_3 > C_2 $) to boost the model capacity.}
    \label{fig:framework}
\end{figure*}

\begin{figure}
    \centering
    \begin{subfigure}{0.23\textwidth}
      \includegraphics[width=1\textwidth]{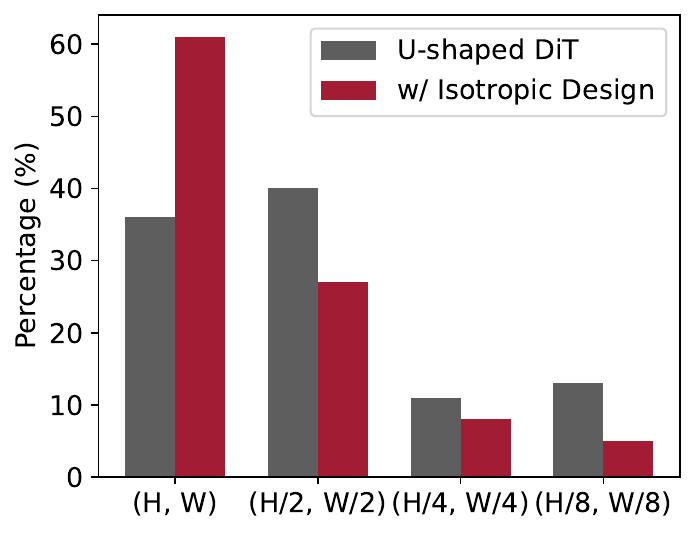}
      \caption{FLOPs}
    \end{subfigure}
    \hfill
    \begin{subfigure}{0.23\textwidth}
      \includegraphics[width=1\textwidth]{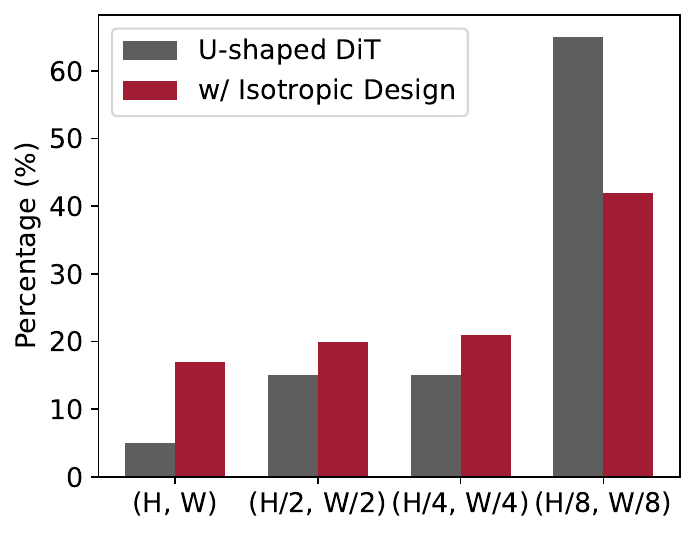}

      \caption{Parameters}

    \end{subfigure}
    \caption{The percentage of FLOPs and parameters for each stage of the U-shaped DiT, both with and without isotropic design, show that more computational resources are allocated to high-resolution stages.}
    \label{fig:reallocate}
\end{figure}

\section{Related Works}
\subsection{Diffusion-based Image Super-Resolution}
Recently, diffusion models~\cite{ho2020denoising,dhariwal2021diffusion} have exhibited substantial benefits in image generation tasks, which generally fall into two categories: train-from-scratch methods and prior-based methods. 
SR3~\cite{saharia2022image} is the pioneer in introducing the diffusion model to the image super-resolution. LDM~\cite{rombach2022high} enhances efficiency by performing the diffusion process in latent space. ResShift~\cite{yue2024resshift} reformulates the diffusion process resulting in a shortened Markov chain that reduces the number of denoising steps to 15. These methods offer significant flexibility and ease of retraining following architectural modifications, making them ideal for lightweight applications.
Inspired by the remarkable potential of Stable Diffusion~\cite{rombach2022high,podell2023sdxl} for text-to-image tasks, several methods~\cite{wang2023exploiting,lin2024diffbir,yang2023pasd,wu2024seesr} have exploited its generative prior to guide real-world image super-resolution. 
While these prior-based methods yield remarkable results, their deployment is limited by slow inference speeds, which arise from the redundant denoiser architecture and the multi-step iterative denoising process.
Despite SinSR~\cite{wang2023sinsr} and AddSR~\cite{xie2024addsr} employing knowledge distillation for one-step denoising, their diffusion architectures typically cannot be altered without massive retraining.
Orthogonal to the efforts to reduce denoising steps, we concentrate on developing an effective diffusion architecture trained from scratch while rivaling the performance of prior-based methods.

\subsection{Diffusion Model Architecture}
Previous diffusion studies~\cite{ho2020denoising, song2020denoising, dhariwal2021diffusion, nichol2021improved, rombach2022high} have predominantly utilized the U-Net~\cite{ronneberger2015u} architecture for denoising, incorporating elements such as ResBlocks~\cite{he2016deep} and Transformer blocks~\cite{vaswani2017attention}.  DiT~\cite{peebles2023scalable} marks a departure from the U-shaped design by adopting an isotropic full transformer architecture, which showcases enhanced scalability. Subsequent works~\cite{ma2024sit,gao2023mdtv2,lu2024fit,li2024hunyuan,gao2024lumina,hatamizadeh2023diffit} have adopted the standard DiT architecture and shown superior performance across various tasks.
U-ViT~\cite{bao2023all} retains the long skip connections typical of U-Net but does not include upsampling or downsampling operations. 
Our proposed DiT architecture, which merges U-shaped and isotropic designs, achieves remarkable performance on image super-resolution.

\section{Preliminaries}
\subsection{Diffusion Models} 
Given a LR image $\bm{y}$ and its corresponding HR image $\bm{x}_0$, diffusion-based SR methods strive to model the conditional distribution $q\left( \bm{x}_0 | \bm{y} \right)$. Typically, these methods define a $T$-step forward process that gradually introduce random noise to $\bm{x}_0$, which can be succinctly achieved in one step through the reparameterization trick:
\begin{equation}
    q\left(\bm{x_t}|\bm{x_0}\right) = \mathcal{N}\left(\bm{x_t} ; \sqrt{\bar{\alpha}_t} \bm{x_0},\left(1-\bar{\alpha}_t\right)\bm{I}\right)  \text { with }  \bar{\alpha}_t=\prod_{i=0}^t \alpha_i,
\end{equation}
where $x_t$ denotes the noised image at time-step $t$ and $\alpha_t$ is the predefined variance schedule. During the reverse process, the model starts from pure Gaussian noise and iteratively generates the preceding state $x_{t-1}$ from $x_t$ using the approximated posterior distribution:
\begin{equation}
    p_\theta \left(\bm{x_{t-1}}|\bm{x_t}, \bm{y_0}\right) = \mathcal{N}\left( \bm{\mu_\theta}\left( \bm{x_t}, \bm{y_0}, t \right), \Sigma\left( \bm{x_t}, t\right)   \right),
\end{equation}
where $\Sigma\left( x_t, t\right)$ is a constant that depends on $\alpha_t$, and $\mu_\theta\left( x_t, y_0, t \right)$ is parameterized by a denoiser $\epsilon_\theta\left( x_t, y_0, t \right)$.

\subsection{Residual Shifting} 
ResShift~\cite{yue2024resshift} constructs a Markov chain between HR and LR images rather than pure Gaussian noise. Let $\bm{e}_0 = \bm{y}_0 - \bm{x}_0$ represents the resduial between the LR and HR images. Additionally, a shifting sequence $\{\eta_t\}_{t=1}^T$ is introduced, gradually increasing from $\eta_1 \to 0$ to $\eta_T \to 1$ with each timeste. The forward process is then formulated based on this shifting sequence:
\begin{equation}
    q(\bm{x}_t|\bm{x}_0, \bm{y}_0) = \mathcal{N}(\bm{x}_t; \bm{x}_0+\eta_t \bm{e}_0, \kappa^2 \eta_t \bm{I}), ~ t=1,2,\cdots,T,
    \label{eq:transit_0_t}
\end{equation}
where $\alpha_t=\eta_t-\eta_{t-1}$ for $t>1$ and $\alpha_1=\eta_1$. The hyper-parameter $\kappa$ controls the noise variance. 
The denoising process, $q(\bm{x}_{t-1}|\bm{x}_t,\bm{x}_0,\bm{y}_0)$  is formulated as follows:
\begin{equation}
    \begin{aligned}
    &p_{\bm{\theta}}(\bm{x}_{t-1}|\bm{x}_t,\bm{x}_0,\bm{y}_0)=\\ 
    &\mathcal{N}\left(\bm{x}_{t-1}\bigg\vert\frac{\eta_{t-1}}{\eta_t}\bm{x}_t+\frac{\alpha_t}{\eta_t}f_{\bm{\theta}}(\bm{x}_t,\bm{y}_0,t), 
    \kappa^2\frac{\eta_{t-1}}{\eta_t}\alpha_t\bm{I}\right),
    \end{aligned}
    \label{eq:poster_elbo}
\end{equation}
where $\bm{x}_0$ is directly predicted by the denoiser $f_{\bm{\theta}}(\bm{x}_t,\bm{y}_0,t)$. This well-designed transfer distribution for image super-resolution effectively reduces the length of Markov chains, thereby reducing the number of required time steps. We follow this paradigm to train our diffusion model.

\begin{figure}[tbp]
    \centering
    \includegraphics[width=1\linewidth]{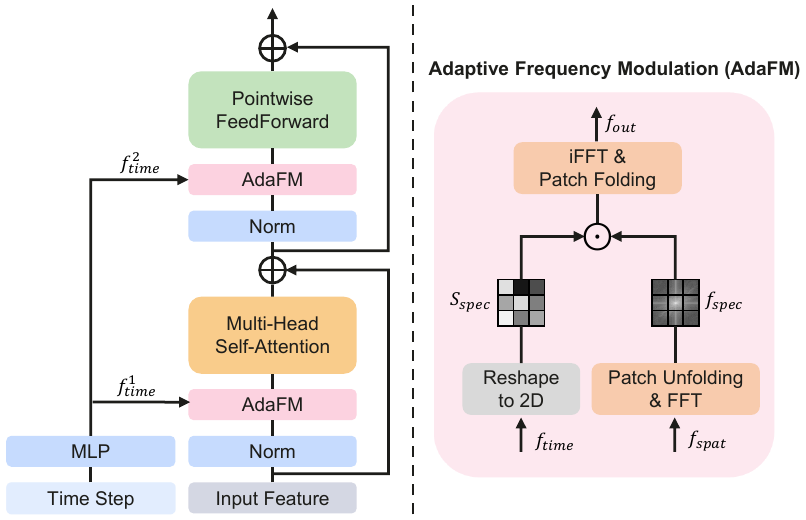}
    \caption{The illustration of transformer block in DiT-SR and Adaptive Frequency Modulation~(AdaFM). AdaFM injects the time step into the frequency domain and adaptively reweights different frequency components.}
    \label{fig:block}
\end{figure}

\section{Methodology}
\subsection{Overall Architecture}
The proposed DiT-SR, depicted in Fig.~\ref{fig:framework}, aims to be trained from scratch to potentially rival the performance of prior-based methods. This denoiser architecture features a U-shaped encoder-decoder global framework but with an isotropic design for all the transformer blocks at different stages. It includes several transformer stages in both the encoder and decoder, each with varying feature resolutions. Within each stage, multiple transformer blocks with uniform configurations are employed, reallocating computational resources to high-resolution layers to enhance the transformer architecture's capacity.

The LR image $\bm{y}$ and noisy image $\bm{x}_t$ are concatenated along the channel dimension, and together with the time step $t$, serve as inputs to the denoiser, which predicts $\bm{x}_0$ and iteratively refines it as outlined in Eq.~\ref{eq:poster_elbo}.
As shown in Fig.~\ref{fig:block}, the transformer block consists of a multi-head self-attention (MHSA) mechanism~\cite{liu2021swin} that operates as a spatial mixer, and a multi-layer perceptron (MLP) with two fully-connected layers separated by GELU activation, serving as channel mixers. Considering the high computational cost and memory constraints of global self-attention when processing high-resolution inputs, we employ local attention with window shifting as an alternative to the original self-attention~\cite{vaswani2017attention}. Group normalization layers are applied before both the MHSA and MLP. Additionally, the proposed Adaptive Frequency Modulation~(AdaFM) is integrated following each normalization layer to inject the time step. Our transformer block can be formulated as:
\begin{equation}
    \begin{array}{l}
        f_{time}^1, f_{time}^2 = \operatorname{MLP_t}(t), \\
        X = \operatorname{MHSA}(\operatorname{AdaFM}(\operatorname{Norm}(X), f_{time}^1)) + X, \\
        X = \operatorname{MLP}(\operatorname{AdaFM}(\operatorname{Norm}(X), f_{time}^2)) + X. \\
    \end{array}
\end{equation}
Subsequent sections will elaborate on the design motivation and specific details of DiT-SR, including the integration of U-shaped global architecture and isotropic block design, as well as the frequency-adaptive time-step conditioning mechanism.

\begin{figure}[tbp]
    \centering
    \includegraphics[width=1\linewidth]{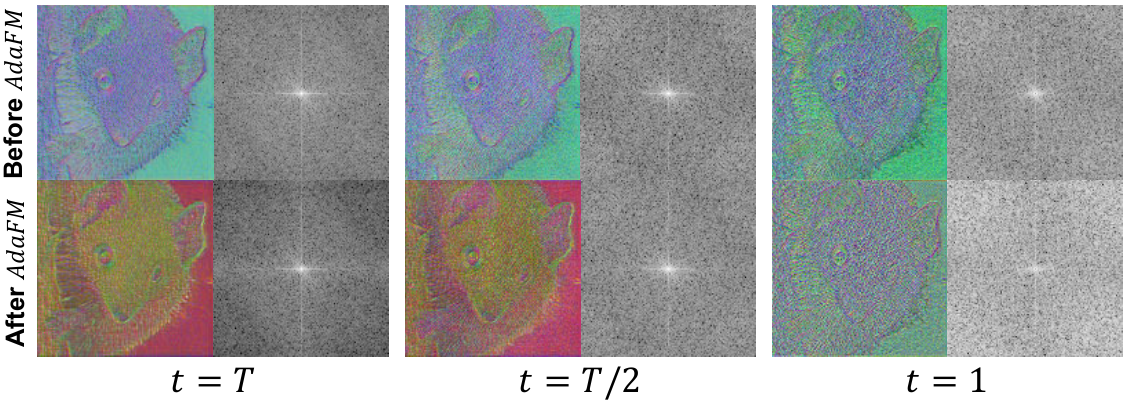}
    \caption{Visualization of the feature maps and their corresponding spectrums before and after applying AdaFM. AdaFM enhances the low-frequency components in the early stages of denoising~(peripheral part of spectrums getting darker) and the high-frequency components in the later stages~(peripheral part of spectrums getting brighter), thereby augmenting the diffusion model's ability to emphasize specific frequency at different time steps.}
    \label{fig:vis_AdaFM}
\end{figure}

\begin{figure*}[t]
    \centering
    \includegraphics[width=1\linewidth]{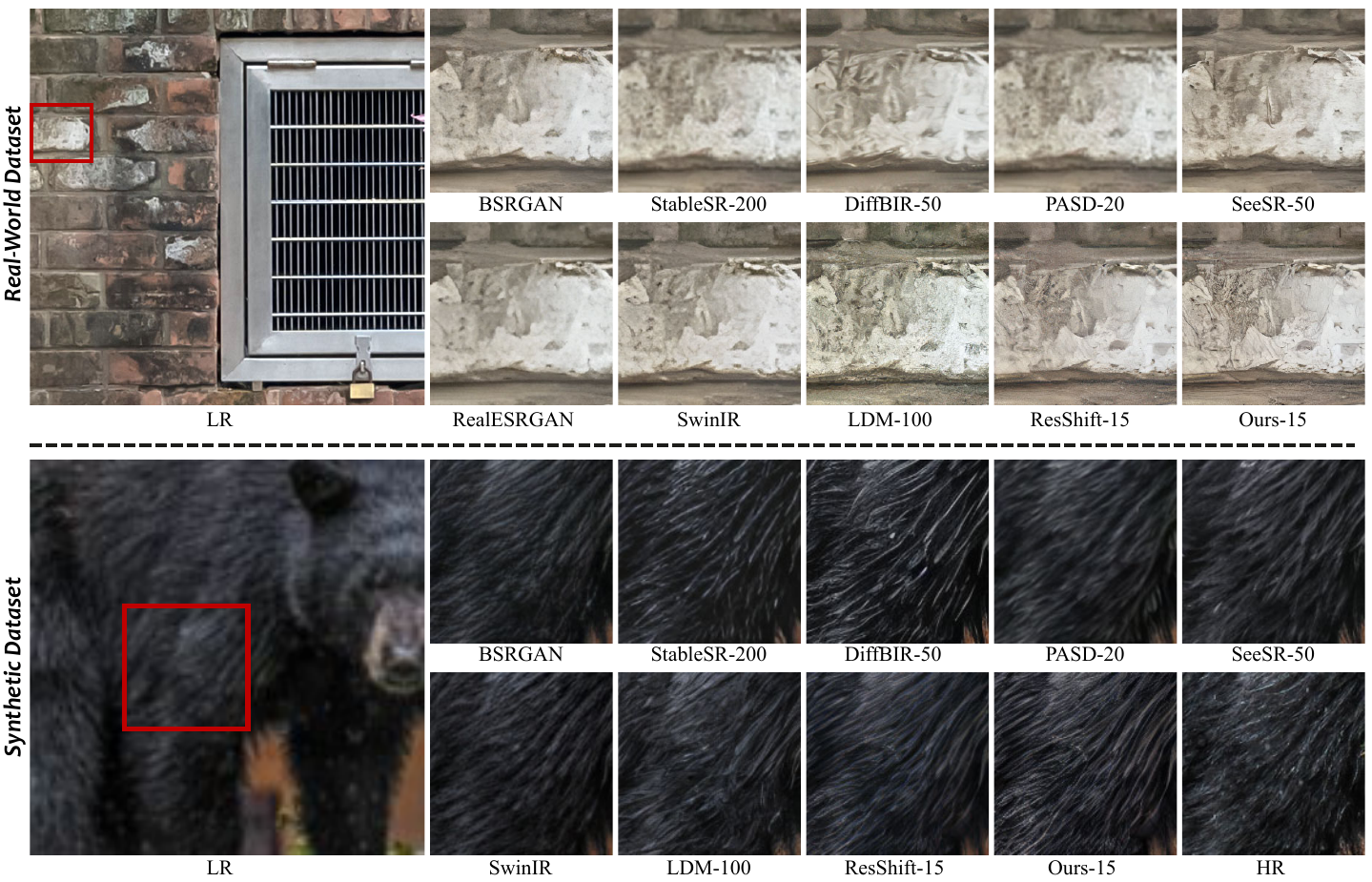}
    \caption{Qualitative comparisons of different methods on both synthetic and real-world datasets. }
    \label{fig:comparision}
\end{figure*}

\subsection{Isotropic Design in U-shaped DiT}
The U-Net architecture~\cite{ronneberger2015u}, with its encoder-decoder framework, is a popular choice for image generation and restoration tasks.  
Given the U-shaped architecture's multi-scale feature extraction ability, we propose integrating the U-shaped global architecture into standard DiT to enhance its performance in image super-resolution. 
The encoder progressively reduces the resolutions of feature maps while increasing their channel dimensions, and the decoder reverses these operations for reconstruction. 

We rethink the isotropic design in DiT and identify two notable characteristics. Firstly, DiTs with consistent channel and depth could effectively handle input with varying patch sizes~(\textit{e.g.}, DiT-XL/2, DiT-XL/4, and DiT-XL/8), which is analogous to processing different resolutions in U-Net. Secondly, DiTs at higher resolutions~(\textit{e.g.}, DiT/2) benefit more from scaling up compared to those at lower resolutions~(\textit{e.g.}, DiT/8). 
Motivated by these insights, we introduce this straightforward yet effective isotropic design to multi-scale U-shaped DiT in a pioneering way. Specifically, each transformer stage consists of several transformer blocks that operate at the same resolution. Within each stage, we standardize the inside feature's channel dimension to be the same across all stages, perform all the transformer blocks in reallocated feature space, and then reassemble them to their original dimensions. 
Considering that high-resolution stages capture more high-frequency details, which are crucial for image super-resolution and exhibit better scalability, we set the standardized channel dimension larger than the original high-resolution stages in U-Net, yet considerably smaller than  low-resolution stages.
As depicted in Fig.~\ref{fig:reallocate}, this isotropic principle allocates computational resources to critical high-resolution layers, avoiding the design of tedious scheduling policies, and greatly boosting the capacity of transformer architecture with far fewer parameters than conventional U-Net.

\begin{table*}[t]
    \renewcommand{\arraystretch}{1.08}
    \centering
    \caption{Performance and denoiser complexity comparison on real-world datasets. The best and second best results are highlighted in \textbf{bold} and \underline{underline}. We denote the number of sampling steps for each diffusion-based method using the format ``method-steps''.}
    \label{tab:real}
    \begin{tabular}{c|c|ccc|ccc}
        \Xhline{0.8pt}
        \multirow{2}{*}{Methods}  & \multirow{2}{*}{$\#$Params} & \multicolumn{3}{c|}{RealSR}  & \multicolumn{3}{c}{RealSet65} \\ 
    \Xcline{3-8}{0.4pt}
        & & CLIPIQA$\uparrow$ & MUSIQ$\uparrow$ & \multicolumn{1}{c|}{MANIQA$\uparrow$} & CLIPIQA$\uparrow$ & MUSIQ$\uparrow$ & MANIQA$\uparrow$ \\ 
        \Xhline{0.4pt}
        \multicolumn{7}{l}{\small \textit{\color{gray}GAN based Methods}} \\
        \Xhline{0.4pt}
        \color{black}RealSR-JPEG         & 17M & 0.3611 & 36.068  & 0.1772 & 0.5278 & 50.5394 & 0.2943    	\\
        \color{black}BSRGAN              & 17M & \color{black}0.5438 & \color{black}63.5819 & \color{black}0.3685 & \color{black}0.616	 & \color{black}65.5774 & \color{black}0.3897	   	\\
        \color{black}RealESRGAN          & 17M & 0.4898 & 59.6766 & 0.3679	& 0.5987 & 63.2228 & 0.3871	   	\\
        \color{black}SwinIR              & 12M & 0.4653 & 59.6316 & 0.3454	& 0.5778 & 63.8212 & 0.3816	   	\\
        \Xhline{0.4pt}
        \multicolumn{7}{l}{\small \color{gray}\textit{Prior based Methods}} \\
        \Xhline{0.4pt}
        \color{black}StableSR-200         & 919M& 0.5207 & 59.4264 & 0.3563 & 0.5338 & 56.9207 & 0.3387	   	\\
        \color{black}DiffBIR-50          & 1670M & \underline{0.7142} & \textbf{66.843} & 0.4802 & \color{black}\textbf{0.7398} & \color{black}\underline{69.7260} & \underline{0.5000}    \\
        \color{black}PASD-20 			 & 1469M & 0.5170 & 58.4394 & 0.3682 & 0.5731 & 61.8813 & 0.3893    	\\
        \color{black}SeeSR-50			 & 1619M & 0.6819 & \underline{66.3461} & \color{black}\textbf{0.5035} & 0.7030 & \textbf{68.9803} & \color{black}\textbf{0.5084}		\\
        \Xhline{0.4pt}
        \multicolumn{7}{l}{\small \color{gray}\textit{Training-from-Scratch Diff. based Methods}} \\
        \Xhline{0.4pt}
        \color{black}LDM-100              & 114M & 0.5969 & 55.4359 & 0.3071 & 0.5936 & 56.112 & 0.356   	\\
        \color{black}ResShift-15         & 119M & 0.6028 & 58.8790 & 0.3891 & 0.6376 & 58.0400 & 0.4048    	\\
        \color{black}Ours-15 & 61M & \color{black}{\textbf{0.7161}} & \color{black}65.8334 & \color{black}{\underline{0.5022}} & \underline{0.7120} & \color{black}{66.7413} & \color{black}{0.4821}   \\
    \Xhline{0.8pt}
    \end{tabular}
\end{table*}

\subsection{Frequency-Adaptive Time Step Conditioning}
Since the diffusion model utilizes the same denoiser across various time steps, it is crucial to explicitly incorporate the time step as a condition. 
Adaptive Layer Normalization (AdaLN), first introduced in DiT~\cite{peebles2023scalable}, has been proven effective on image generation and is widely adopted in subsequent DiT-based models. Nevertheless, unlike image generation tasks, which start from pure noise and focuse primarily on semantics, the SR task emphasizes the recovery of high-frequency details, necessitating the diffusion model to possess strong frequency perception capabilities. 

Our investigation into the temporal evolution of images predicted by the diffusion-based super-resolution model reveals that it focuses on various frequency components at different denoising stages. As shown in Fig.~\ref{fig:sr_freq}, the model initially reconstructs low-frequency elements, corresponding to the image structure, and progressively refines high-frequency details, associated with texture. Consequently, the time step should adaptively modulate different frequency components, using distinct modulation parameters for high and low-frequency regions.
However, AdaLN modulates feature maps exclusively in the channel dimension, applying uniform modulation parameters across all spatial locations. This limitation hinders its ability to effectively address the specific frequency requirements of image super-resolution tasks. Moreover, it is challenging to generate modulation spatial-wise parameters from a one-dimensional time-step vector, as it requires adaptively distinguishing between the high and low-frequency components' spatial positions in the input image.

To solve this challenge, we introduce Adaptive Frequency Modulation~(AdaFM), replacing AdaLN after each normalization layer and switching the time-step modulation from the spatial domain to the frequency domain, as shown in Fig.~\ref{fig:block}.  Initially, to accommodate various input resolutions and enhance efficiency, we segment the spatial domain feature map $f_{spat} \in \mathbb{R}^{C \times H \times W}$ into $p\times p$ windows. Subsequently, we transform these segments into spectrograms $f_{spec} \in \mathbb{R}^{\frac{H \times W}{p^2} \times C \times p \times p}$  using the Fast Fourier Transform within each window. The time step is mapped to a $p^2$-dimensional vector $f_{time}$ and reshaped into a frequency scale matrix $S_{spec} \in \mathbb{R}^{p \times p}$, which is then used to adaptively reweight various frequency components, thereby augmenting the diffusion model's ability to emphasize specific frequency at different time steps, as illustrated in Fig.~\ref{fig:vis_AdaFM}. 

In a spectrum, each pixel at a specific spatial position corresponds to a predetermined frequency component, defined solely by the feature map's spatial dimension, independent of its content. The frequency corresponding to a pixel located at spatial position $(u, v)$ in spectrum $\in \mathbb{R}^{H \times W}$ can be formulated as:
\begin{equation}
    f_{u} = \frac{u - H / 2}{H} \times F_s, \quad f_{v} = \frac{v-W/2}{H}  \times F_s,
\end{equation}
where $f_u$, $f_v$ denote the vertical and horizontal frequencies separately, and $F_s$ indicates sampling frequency.
This consistency allows the same frequency scale matrix $S_{spec}$ to be applied across all windows and channels, significantly enhancing efficiency. In comparison to AdaLN, which requires $dim_{f_{time}} \times C \times 3 \times 2$ mapping parameters~($scale$, $shift$ and $gate$ for both self-attention and MLPs), AdaFM requires only $dim_{f_{time}} \times p^2 \times 2$.
The process is formulated as follows:
\begin{equation}
    \begin{array}{l}
        S_{spec} = \operatorname{reshape}(f_{time}, p \times p), \\
        f_{spec}=\operatorname{FFT}\left(\mathcal{P}\left(f_{spat}\right)\right), \\
        f^{'}_{spec}=S_{spec} \odot f_{spec}, \\
        f_{out}=\mathcal{P}^{-1}\left(\operatorname{iFFT}\left(f^{'}_{spec}\right)\right),
    \end{array}
\end{equation}
where $\mathcal{P}$ and $\mathcal{P}^{-1}$ denote the patch unfolding and folding operations, $\operatorname{FFT}$ and $\operatorname{iFFT}$ indicate Fast Fourier Transform and inverse  Fourier Transform. Given that different frequencies correspond to distinct spatial locations on the feature map, the proposed frequency-wise time-step conditioning module actually provides spatial-wise modulation.

\begin{table*}[t]
    \centering
    \caption{Ablation Study on real-world datasets. The percentage reductions in the number of parameters and FLOPs are compared to the U-shaped DiT. The best results are highlighted in \textbf{bold}.}
    \label{tab:ablation}
    \begin{tabular}{cc|cc|cc|cc}
        \Xhline{0.8pt}
        \multicolumn{2}{c|}{Configuration} & \multirow{2}{*}{$\#$Params} & \multirow{2}{*}{FLOPs}                              
        & \multicolumn{2}{c|}{RealSR}  & \multicolumn{2}{c}{RealSet65} \\ 
        \Xcline{5-8}{0.4pt}
        DiT Arch. & Time Conditioning & & & CLIPIQA$\uparrow$ & MUSIQ$\uparrow$ & CLIPIQA$\uparrow$ & MUSIQ$\uparrow$ \\ 
        \Xhline{0.4pt}
        Isotropic &	AdaLN	& 42.38M  & 122.99G & 0.655 & 64.194 & 0.664 & 64.263 			\\
        U-shape   & AdaLN   & 264.39M & 122.87G & 0.688 & 64.062 & 0.693 & 65.604 	\\
        Ours & AdaLN   & 100.64M\color{lightblue}(-62\%) & 93.11G\color{lightblue}(-24\%)  & 0.700 & 64.676 & 0.699 & \textbf{67.634} \\
        Ours &	AdaFM	& 60.79M\color{lightblue}(-77\%) & 93.03G\color{lightblue}(-24\%)  & \textbf{0.716} & \textbf{65.833} & \textbf{0.712} & 66.741	\\
        \Xhline{0.8pt}
    \end{tabular}
\end{table*}

\section{Experiments}
\subsection{Experimental Settings}
\subsubsection{Datasets}
We evaluate the proposed model on $\times 4$ real-world SR task. The training data comprises LSDIR~\cite{li2023lsdir}, DIV2K~\cite{Agustsson_2017_CVPR_Workshops}, DIV8K~\cite{gu2019div8k}, OutdoorSceneTraining~\cite{wang2018recovering}, Flicker2K~\cite{timofte2017ntire} and the first 10K face images from FFHQ~\cite{karras2019style} datasets. We partition LSDIR into a training set with 82991 images and a test set with 2000 images. Following LDM~\cite{rombach2022high}, HR images in our training set are randomly cropped to $256\times256$ and the degradation pipeline of RealESRGAN~\cite{wang2021real} is used to synthesize LR/HR pairs. The test set images are center-cropped to $512\times512$ and subjected to the same degradation pipeline used in the training stage to create a synthetic dataset, named LSDIR-Test. Furthermore, we utilize two real-world datasets: RealSR~\cite{cai2019toward}, which comprises 100 real images captured by Canon 5D3 and Nikon D810 cameras, and RealSet65~\cite{yue2024resshift}, including 65 low-resolution images collected from widely used datasets and the internet.

\subsubsection{Implementation Details}
Following LDM~\cite{rombach2022high}, the proposed architecture operates in latent space, utilizing the Vector Quantized GAN (VQGAN)~\cite{esser2021taming} with a downsampling factor of 4.
We train the proposed model for $300K$ iterations with a batch size of $64$ using 8 NVIDIA Tesla V100 GPUs. The optimizer is Adam~\cite{kingma2014adam}, and the learning rate is $5e^{-5}$. The FFT window size $p$ is empirically set to $8$~\cite{wallace1991jpeg,kong2023efficient}. Detailed architectural configurations are provided in the supplementary material.

\subsubsection{Evaluation Metrics}
We adopt reference-based metrics, including PSNR and LPIPS~\cite{zhang2018unreasonable}, to evaluate the performance of different models. Additionally, non-reference metrics such as CLIPIQA~\cite{wang2023exploring}, MUSIQ~\cite{ke2021musiq}, and MANIQA~\cite{yang2022maniqa}, which are more consistent with human perception in generative SR, are also employed. For assessments on real-world datasets, due to the lack of ground truth, we evaluate their performance using only non-reference metrics. 

\subsection{Comparison with State-of-the-Arts}
\subsubsection{Comparison Methods}
We compared our proposed architecture with several latest SR methods, including
GAN-based methods such as RealSR-JPEG\cite{ji2020real}, BSRGAN\cite{zhang2021designing}, RealESRGAN\cite{wang2021real}, and SwinIR\cite{liang2021swinir}, as well as diffusion-based methods like LDM\cite{rombach2022high}, StableSR\cite{wang2023exploiting}, ResShift\cite{yue2024resshift}, DiffBIR~\cite{lin2024diffbir}, PASD~\cite{yang2023pasd} and ~\cite{wu2024seesr}. The steps are configured using their default settings.
It is worth noting that StableSR, DiffBIR, PASD and SeeSR leverage the generative prior of Stable Diffusion, which is pretrained on large-scale datasets for thousands of GPU days, while LDM and ResShift are trained from scratch like ours.

\subsubsection{Comparison on Real-World and Synthetic Datasets}
We present the qualitative and quantitative results on Fig.~\ref{fig:comparision}, Tab.~\ref{tab:real} and Tab.~\ref{tab:lsdir}. The proposed architecture significantly outperforms existing training-from-scratch methods, and even surpasses  competitive with state-of-the-art prior-based methods, while utilizing only about $ 5\%$ of their parameters. 

\begin{table}[t]
    \small 
    \centering
    \setlength{\tabcolsep}{1mm}
    \caption{Performance comparison on the synthetic LSDIR-Test dataset. The best and second best results are highlighted in \textbf{bold} and \underline{underline}.}
    \label{tab:lsdir}
    {
        \begin{tabular}{@{}C{1.8cm}|@{}C{1.1cm}@{} @{}C{1.1cm}@{} @{}C{1.2cm}@{} @{}C{1.2cm}@{}c}
            \Xhline{0.8pt}
            \multirow{2}{*}{Methods} & \multicolumn{5}{c}{LSDIR-Test} \\ 
        \Xcline{2-6}{0.4pt}
            & PSNR$\uparrow$ & LPIPS$\downarrow$ & \multicolumn{1}{l}{CLIPIQA$\uparrow$} & MUSIQ$\uparrow$ & MANIQA$\uparrow$ \\ \hline
            \multicolumn{5}{l}{\small \color{gray}\textit{GAN based Methods}} \\
            \Xhline{0.4pt}
            RealSR-JPEG       & 22.16 & 0.360  & 0.546  & 59.02 & 0.342 \\
            BSRGAN            & \underline{23.74} & 0.274 & 0.570 & 67.94 & 0.394 \\
            RealESRGAN        & 23.15 & 0.259 & 0.568 & 68.23 & 0.414	\\
            SwinIR        & 23.17 & \underline{0.247}  & 0.598  & 68.20 & 0.414	\\
        \Xhline{0.4pt}
        \multicolumn{5}{l}{\small \color{gray}\textit{Prior based Methods}} \\
            \Xhline{0.4pt}
            StableSR-200       & 22.68 & 0.267  & 0.660 & 68.91 & 0.416	\\
                DiffBIR-50        & 22.84 & 0.274  & \underline{0.709} & \underline{70.05} & 0.455	  \\
                PASD-20           & 23.57 & 0.279  & 0.624 &  69.07 & 0.440 	 \\
            SeeSR-50 		  & 22.90 & 0.251  & \textbf{0.718} & \textbf{72.47} & \textbf{0.559}	 \\
        \Xhline{0.4pt}
        \multicolumn{5}{l}{\small \color{gray}\textit{Training-from-Scratch Diff. based Methods}}\\
        \Xhline{0.4pt}
                LDM-100            & 23.34 & 0.255  & 0.601 & 66.84 & 0.413	  \\
                ResShift-15       & \textbf{23.83} & \underline{0.247}  & 0.640	 & 67.74 & 0.464 \\ 
            Ours-15   & 23.60 & \textbf{0.244} & 0.646 & 69.32 & \underline{0.483}	\\ 
        \Xhline{0.8pt}
        \end{tabular}
    }
\end{table}

\subsection{Ablation Study}
\subsubsection{U-shaped DiT with Isotropic Design}
As described above, this paper proposes an evolutionary path from standard DiT to U-shaped DiT, and ultimately introduces isotropic design to multi-scale U-shaped DiT.  We reimplement DiT for super-resolution, employing local attention with window shifting to replace the original self-attention. As shown in Table~\ref{tab:ablation}, the U-shaped DiT outperforms the standard DiT for the same FLOPs, but has six times more parameters. 
Notably, by reallocating computational resources to critical layers within the isotropic design, performance is improved even with a 62\% reduction in parameters.

\subsubsection{Adaptive-Frequency Modulation}
The proposed AdaFM operates in the frequency domain, adaptively identifying high and low-frequency regions and modulating them separately with distinct parameters. Additionally, due to the nature of the frequency domain and our highly efficient design, the parameter count of AdaFM is only a fraction of that of AdaLN.
As shown in Tab.~\ref{tab:ablation}, replacing AdaLN with AdaFM reduced the number of denoiser parameters from 100.64M to 60.79M, while also enhancing model performance, demonstrating the effectiveness of AdaFM. Fig.~\ref{fig:vis_AdaFM} visualizes the feature maps and their spectrums before and after AdaFM, illustrating how it adaptively enhances low-frequency components in the early stages of denoising and high-frequency components in the later stages, thereby establishing a correlation between the time step and frequency.

\section{Discussion and Conclusion}
In this work, we introduce DiT-SR, an effective diffusion transformer architecture for image super-resolution that can be trained from scratch to rival the performance of prior-based methods. It integrates U-shaped global architecture and isotropic block designs, reallocating the computational resources to critical high-resolution layers, and boosting the performance efficiently. 
Furthermore, we propose an efficient yet effective time-step conditioning module AdaFM that adaptively reweights different frequency components, augmenting the diffusion model's ability to emphasize specific frequency information at varying time steps. 

\noindent \textbf{Future Work.}
AdaFM holds the potential to establish a new time-step conditioning paradigm for diffusion models, extending its application to various low-level visual tasks and even to text-to-image generation that also adheres to the paradigm of initially generating low frequencies followed by high frequencies.

\noindent \textbf{Limitation and Ethical Statement.}
Due to differences in tasks and limited data, image super-resolution models typically do not exhibit the same level of scalability as text-to-image models. Although our denoiser achieves competitive performance with much fewer parameters compared to prior-based models, it still has some way to go before fully surpassing their performance. Similar to other content generation methods, our approach must be used cautiously to prevent potential misuse.

{
    \small
    \bibliographystyle{ieeenat_fullname}
    \bibliography{main}
}

\appendix

\begin{table*}[!h]
	\centering
	\caption{Diffusion Architecture Hyper-parameters.}
	\label{tab:arch_detail}
	\begin{tabular}{ccccccc}
		\Xhline{0.8pt}
		DiT Arch. & Time Conditioning & $\#$Params & FLOPs & Number of Blocks & Channels & Reallocated Channel \\ 
		\Xhline{0.4pt}
		Isotropic & AdaLN & 42.38M  & 122.99G & [6,6,6,6,6] & 160 & -		\\
		U-shape   & AdaLN & 264.39M & 122.87G & [6,6,6,6]   & [160,320,320,640] & -\\
		Ours      & AdaLN & 100.64M & 93.11G  & [6,6,6,6]   & [160,320,320,640] & 192 \\
		Ours      & AdaFM & 60.79M  & 93.03G  & [6,6,6,6]   & [160,320,320,640] & 192 \\
		Ours-Lite & AdaFM & 30.89M  & 49.17G  & [4,4,4]	    & [128,256,256] 	& 160 \\
		\Xhline{0.8pt}
	\end{tabular}
\end{table*}

\section{Architecture Details}
In the main paper, we present several diffusion architectures, including the standard DiT with an isotropic design, the U-shaped DiT, and our proposed architecture, which combines a U-shaped global structure with an isotropic block design. 
For the standard DiT, we re-implemented it inspired by SwinIR's Residual Swin Transformer Block (RSTB)~\cite{liang2021swinir}, where each stage includes a residual connection. In this architecture, the feature resolution and channel dimensions remain consistent across all stages.
The U-shaped DiT incorporates spatial downsampling and upsampling, with channel dimensions increasing progressively as depth increases. This design choice results in a much higher parameter count for the U-shaped DiT compared to the standard DiT, even when both architectures have the same FLOPs.
Our proposed architecture strategically expands channels in the high-resolution layers and compresses them in the low-resolution layers. By reallocating limited parameters to the most critical layers, this design maximizes performance, surpassing even the original model. This approach effectively balances computational resources and enhances the model's ability to capture and refine crucial high-frequency details, which are essential for tasks like image super-resolution.
The transformer block number is set to $6$ for each stage, and the base channel is configured to $160$. For the U-shaped global architecture, there are 4 stages, with the channel increase factor set to $[1,2,2,4]$. The specific architectural details are provided in Tab.~\ref{tab:arch_detail}. 

\section{Compressing the U-shaped DiT}
As described in the main paper, our proposed architecture outperforms the U-shaped DiT with only $48\%$ of the parameters.  This surprising result raises the question of whether U-Net architectures might be overly redundant, potentially limiting their performance. To explore this, we applied two separate compression strategies to the U-shaped DiT, adjusting its depth and width.
For depth compression, we reduced the number of transformer blocks from $6$ to $4$ per stage, creating a "Narrower U-DiT." For width compression, we decreased the base channel dimension from $160$ to $144$. Despite these relatively modest reductions in parameter count (approximately $20\%$), both strategies resulted in noticeable performance degradation, as shown in Tab.~\ref{tab:compressing}. This outcome indicates that U-Net architecture with $d6c160$ configuration has not yet reached a point of redundancy. Furthermore, it supports the effectiveness of our approach, which reallocates computational resources to critical high-resolution layers, thereby achieving superior performance with fewer parameters.

\begin{table*}[t]
	\centering
	\caption{The results of compressing U-shaped DiT on real-world datasets. The percentage reductions in the number of parameters and FLOPs are compared to the U-shaped DiT. The best results are highlighted in \textbf{bold}.}
	\label{tab:compressing}
	\begin{tabular}{c|cc|cc|cc}
		\Xhline{0.8pt}
		\multirow{2}{*}{Methods} & \multirow{2}{*}{$\#$Params} & \multirow{2}{*}{FLOPs}                              
		& \multicolumn{2}{c|}{RealSR}  & \multicolumn{2}{c}{RealSet65} \\ 
		\Xcline{4-7}{0.4pt}
		& & & CLIPIQA$\uparrow$ & MUSIQ$\uparrow$ & CLIPIQA$\uparrow$ & MUSIQ$\uparrow$ \\ 
		\Xhline{0.4pt}
		U-shaped DiT        & 264.39M & 122.87G & 0.688 & 64.062 & 0.693 & 65.604 	\\
		Shallower U-DiT		& 196.65M\color{gray}(-26\%) & 96.30G\color{gray}(-22\%)  & 0.671 & 63.319 & 0.683 & 64.097 	\\
		Narrower U-DiT		& 214.20M\color{gray}(-19\%)  & 99.56G\color{gray}(-19\%)  & 0.682 & 63.631 & 0.692 & 65.469	\\
		Ours w/ AdaLN    & 100.64M\color{lightblue}(-62\%) & 93.11G\color{lightblue}(-24\%)  & \textbf{0.700} & \textbf{64.676} & \textbf{0.699} & \textbf{67.634} \\
		\Xhline{0.8pt}
	\end{tabular}
\end{table*}

\begin{table*}[!h]
	\centering
	\caption{The results of compressing U-shaped DiT on real-world datasets.}
	\label{tab:lightweight}
	\begin{tabular}{c|c|ccc|ccc}
		\Xhline{0.8pt}
		\multirow{2}{*}{Methods} & \multirow{2}{*}{$\#$Params} & \multicolumn{3}{c|}{RealSR}  & \multicolumn{3}{c}{RealSet65} \\ 
		\Xcline{3-8}{0.4pt}
		& & CLIPIQA$\uparrow$ & MUSIQ$\uparrow$ & MANIQA$\uparrow$ & CLIPIQA$\uparrow$ & MUSIQ$\uparrow$ & MANIQA$\uparrow$ \\ 
		\Xhline{0.4pt}
		LDM-100      & 114M & 0.5969 & 55.4359 & 0.3071 & 0.5936 & 56.1120 & 0.3560   	\\
		ResShift-15  & 119M & 0.6028 & 58.8790 & 0.3891 & 0.6376 & 58.0400 & 0.4048    	\\
		Ours-15      & 61M  & 0.7161 & 65.8334 & 0.5022 & 0.7120 & 66.7413 & 0.4821  	\\
		Ours-Lite-15 & 31M  & 0.6670 & 63.0544 & 0.4565 & 0.6694 & 64.3387 & 0.4420		\\
		Ours-Lite-1  & 31M  & 0.6993 & 63.3759 & 0.4262 & 0.7092 & 64.8329 & 0.4299 	\\
		\Xhline{0.8pt}
	\end{tabular}
\end{table*}

\section{Lightweight Version}
Different from prior-based super-resolution methods, training-from-scratch methods offer significant flexibility and ease of retraining after architectural modifications. This adaptability makes them particularly well-suited for lightweight applications. In addition to the base version of our proposed model reported in the main paper, we also develop a lite version. As illustrated in Tab.~\ref{tab:arch_detail}, the number of transformer blocks is reduced from $6$ to $4$, and the base channel is reduced from $160$ to $128$ with the channel increase factor $[1,2,2]$. Inspired by BK-SDM\cite{kim2023bksdm}, we also remove the deepest layer to further streamline the model. As shown in Tab.~\ref{tab:lightweight}, even though our lightweight model has only $25\%$ of the parameters compared to ResShift\cite{yue2024resshift}, a state-of-the-art diffusion-based SR method trained from scratch, it still significantly outperforms ResShift, further demonstrating the superior model capacity of our proposed diffusion architecture.

Notably, our method is orthogonal to existing step-distillation methods and can be combined to enhance inference efficiency. Specifically, we utilize SinSR~\cite{wang2023sinsr} to distill the lightweight model from a 15-step denoising process to a single-step denoising process.  After applying step distillation, the CLIPIQA and MUSIQ metrics improved, while the MANIQA metric showed a decline. Users can choose whether to distill the model to achieve single-step denoising based on the specific requirements of their application scenario.

\section{More Visualization Results}
\begin{figure*}[htbp]
	\centering
	\includegraphics[width=1\linewidth]{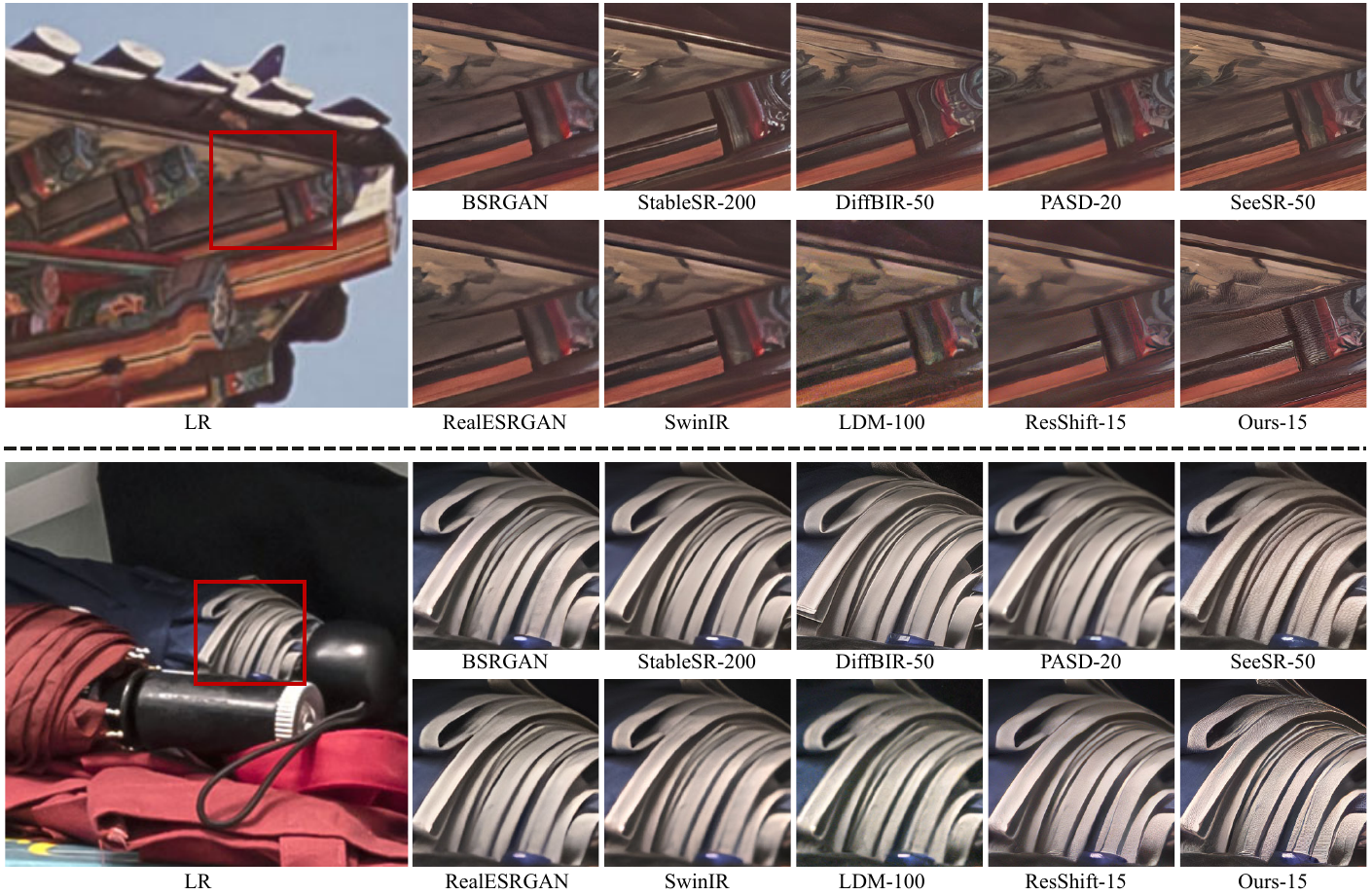}
	\caption{More visualization results on real-world datasets. Please zoom in for a better view.
	}
	\label{fig:more_vis_realworld}
\end{figure*}

We provide more real-world image super-resolution results in Fig.~\ref{fig:more_vis_realworld}.

\section{Experiments on Blind Face Restoration}
\paragraph{Training Details.}
We use FFHQ\cite{karras2019style} dataset containing 70K high-quality (HQ) face images with a resolution of 1024 $ \times $ 1024. Firstly we resized the HQ images into 512 $\times$ 512, and then processed a typical degradation pipeline~\cite{wang2021towards} to synthesize the LQ images.
The learning rate grows to 5e-5 in 5000 iterations, then gradually decays from 5e-5 to 2e-5 according to the annealing cosine schedule and training ends at 200K iterations. 
Following ResShift~\cite{yue2024resshift}, we employ VQGAN~\cite{esser2021taming} with a downsampling factor of 8. The diffusion step is set to $4$. In addition to diffusion loss~\cite{ho2020denoising} in latent space, LPIPS~\cite{zhang2018unreasonable} loss is also adopted in pixel space. 
The model is trained with a batch size of 64 using 8 NVIDIA Tesla V100 GPUs.

\paragraph{Test Datasets.}
We use 2000 HR images that randomly selected from the validation dataset of CelebA-HQ\cite{karras2017progressive} as the test datasets and the corresponding LQ images are synthesized following GFPGAN~\cite{wang2021towards}. Additionally, three typical real-world datasets, named LFW\cite{huang2008labeled}, WebPhoto\cite{wang2021towards}, and WIDER\cite{zhou2022towards} are used to evaluate the performance on different degrees of degradation. 
LFW is a widely used face recognition dataset comprising 1,711 face images collected from various real-world sources. It provides a standard benchmark for evaluating face recognition algorithms under natural, unconstrained conditions.
WebPhoto consists of 407 face images obtained through web crawling. This dataset includes a diverse range of images, including some older photos with significant degradation. It offers a variety of visual content and poses challenges for large-scale image retrieval and clustering algorithms due to its diverse and sometimes low-quality images.
WIDER includes a subset of 970 face images selected from the larger WIDER Face dataset. The subset features images with heavy degradation, including occlusions, variations in poses, scales, and lighting conditions. It serves as a robust benchmark for assessing the effectiveness of face restoration methods under challenging real-world scenarios.

\begin{table*}[htbp]
	\centering
	\caption{Quantitative results of different methods on the dataset of \textit{CelebA-Test}. The best and second best results are highlighted in \textbf{bold} and \underline{underline}.}
	\label{tab:celeba_testing}
	\begin{tabular}{@{}C{3.0cm}@{}|
			@{}C{1.5cm}@{} @{}C{1.5cm}@{} @{}C{1.5cm}@{} @{}C{1.6cm}@{} @{}C{1.6cm}@{} @{}C{1.6cm}@{} @{}C{1.6cm}@{} @{}C{1.6cm}}    
		\Xhline{0.8pt}
		\multirow{2}*{Methods} & \multicolumn{7}{c}{CelebA-Test} \\
		\Xcline{2-8}{0.4pt}
		& LPIPS$\downarrow$ & IDS$\downarrow$ & LMD$\downarrow$ & FID$\downarrow$ & CLIPIQA$\uparrow$ & MUSIQ$\uparrow$ & ManIQA  \\
		\Xhline{0.4pt}
		DFDNet       & 0.739 & 86.323 & 20.784 & 76.118 & 0.619 & 51.173 & 0.433  \\
		PSFRGAN		 & 0.475 & 74.025 & 10.168 & 60.748 & 0.630 & 69.910 & 0.477  \\
		GFPGAN		 & 0.416 & 66.820 & 8.886 & 27.698 & 0.671 & 75.388 & \underline{0.626}  \\
		VQFR      	 & 0.411 & 65.538 & 8.910 & 25.234 & 0.685 & 73.155 & 0.568 \\
		CodeFormer	 & \underline{0.324} & \textbf{59.136} & \textbf{5.035} & 26.160 & \underline{0.698} & \textbf{75.900} & 0.571 \\
		DiffFace-100 & 0.338 & 63.033 & 5.301 & 23.212 & 0.527 & 66.042 & 0.475  \\
		ResShift-4	 & \textbf{0.309} & \underline{59.623} & \underline{5.056} & \textbf{17.564} & 0.613 & 73.214 & 0.541 \\
		Ours-4		 & 0.337 & 61.4644 & 5.235  & \underline{19.648} & \textbf{0.725} & \underline{75.848} & \textbf{0.634}	\\
		\Xhline{0.8pt}
	\end{tabular} 
\end{table*}

\begin{figure*}[htbp]
	\centering
	\includegraphics[width=1\linewidth]{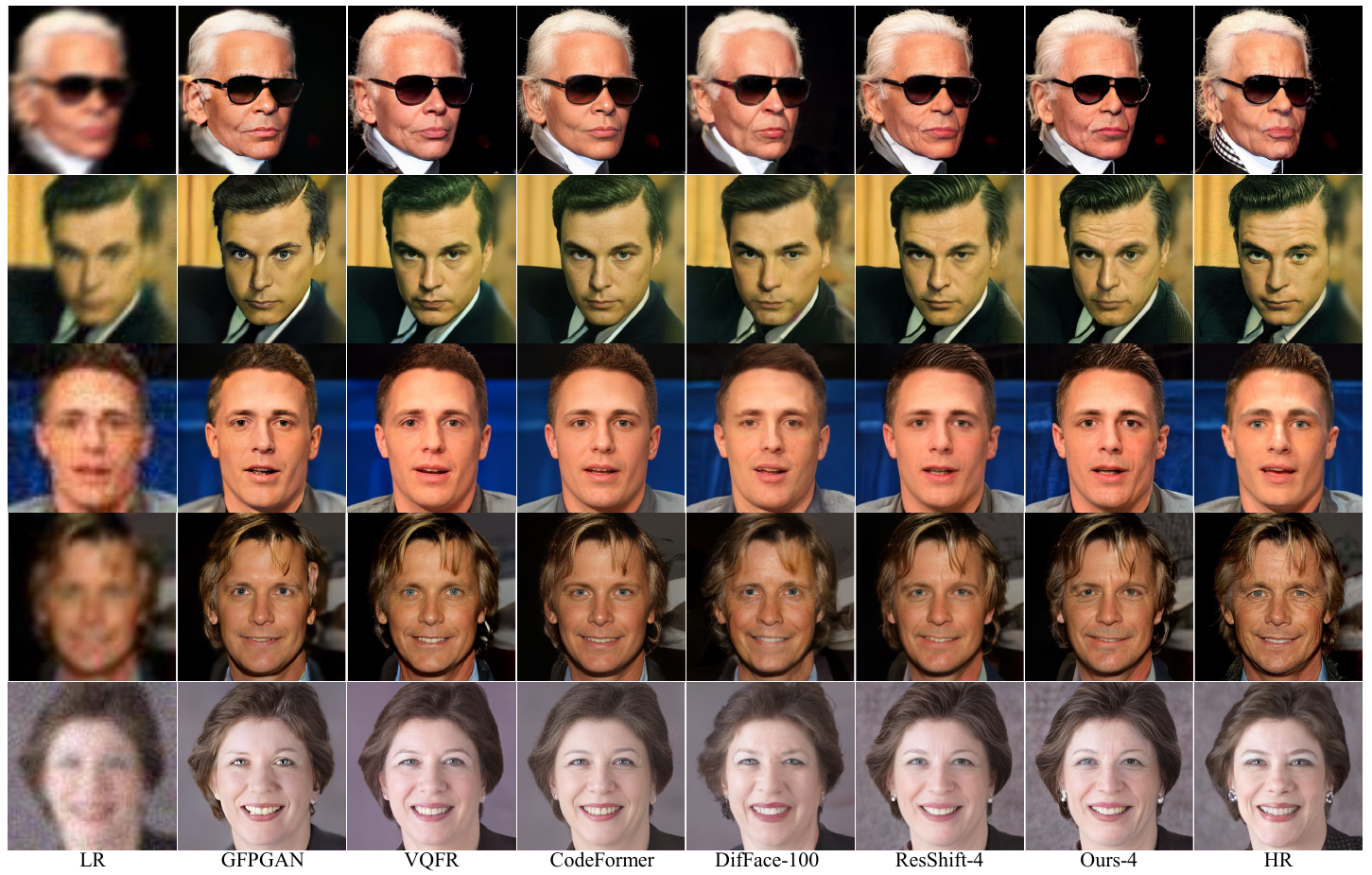}
	\caption{Qualitative results of different methods on synthetic CelebA-Test dataset for blind face restoration. Please zoom in for a better view.
	}
	\label{fig:face_celeb}
\end{figure*}

\begin{table*}[htbp]
	\centering
	\caption{Quantitative results of different methods on three real-world human face datasets. The best and second best results are highlighted in \textbf{bold} and \underline{underline}.}
	\label{tab:real_face_testing}
	\scalebox{1}{
	\begin{tabular}{@{}C{3.0cm}@{}|
			@{}C{1.6cm}@{} @{}C{1.6cm}@{} @{}C{1.6cm}@{}| 
			@{}C{1.6cm}@{} @{}C{1.6cm}@{} @{}C{1.6cm}@{}|
			@{}C{1.6cm}@{} @{}C{1.6cm}@{} @{}C{1.6cm}@{} }
		\Xhline{0.8pt}
		\multirow{2}*{Methods}
		& \multicolumn{3}{c|}{LFW}  & \multicolumn{3}{c|}{WebPhoto} & \multicolumn{3}{c}{Wider} \\
		\Xcline{2-10}{0.4pt}
		& CLIPIQA$\uparrow$ & MUSIQ$\uparrow$  & MANIQA$\uparrow$   
		& CLIPIQA$\uparrow$ & MUSIQ$\uparrow$  & MANIQA$\uparrow$   
		& CLIPIQA$\uparrow$ & MUSIQ$\uparrow$  & MANIQA$\uparrow$   \\
		\Xhline{0.4pt}
		DFDNet     & \underline{0.716} & 73.109 & \textbf{0.6062} & 0.654 & 69.024 & 0.550 & 0.625 & 63.210 & 0.514 \\
		PSFRGAN    & 0.647 & 73.602 & 0.5148 & 0.637 & 71.674 & 0.476 & 0.648 & 71.507 & 0.489 \\
		GFPGAN     & 0.687 & \underline{74.836} & \underline{0.5908} & 0.651 & 73.367 & \textbf{0.577} & 0.663 & \textbf{74.694} & \textbf{0.602} \\
		VQFR       & 0.710 & 74.386 & 0.5488 & 0.677 & 70.904 & 0.511 & \underline{0.707} & 71.411 & 0.520 \\
		CoderFormer & 0.689 & \textbf{75.480} & 0.5394 & \underline{0.692} & \textbf{74.004} & 0.522& 0.699 & 73.404 & 0.510 \\
		DiffFace-100 & 0.593 & 70.362 & 0.4716 & 0.555 & 65.379 & 0.436 & 0.561 & 64.970 & 0.436 \\
		ResShift-4  & 0.626 & 70.643 & 0.4893  & 0.621 & 71.007 & 0.495 & 0.629 & 71.084 & 0.494 \\
		Ours-4	& \textbf{0.727} & 73.187 & 0.564 & \textbf{0.717} & \underline{73.921} & \underline{0.571} & \textbf{0.743} & \underline{74.477} & \underline{0.589}	\\
		\Xhline{0.4pt}
	\end{tabular}} 
	\vspace{-4mm}
\end{table*}

\paragraph{Comparison Methods.}
We compare our method with eight recent blind face restoration methods, including DFDNet~\cite{li2020blind}, PSFRGAN\cite{chen2021progressive}, GFPGAN\cite{wang2021towards}, VQFR\cite{gu2022vqfr}, CodeFormer\cite{zhou2022towards}, DifFace\cite{yue2022difface}, and ResShift\cite{yue2024resshift}.

\begin{figure*}[htbp]
	\centering
	\includegraphics[width=0.9\linewidth]{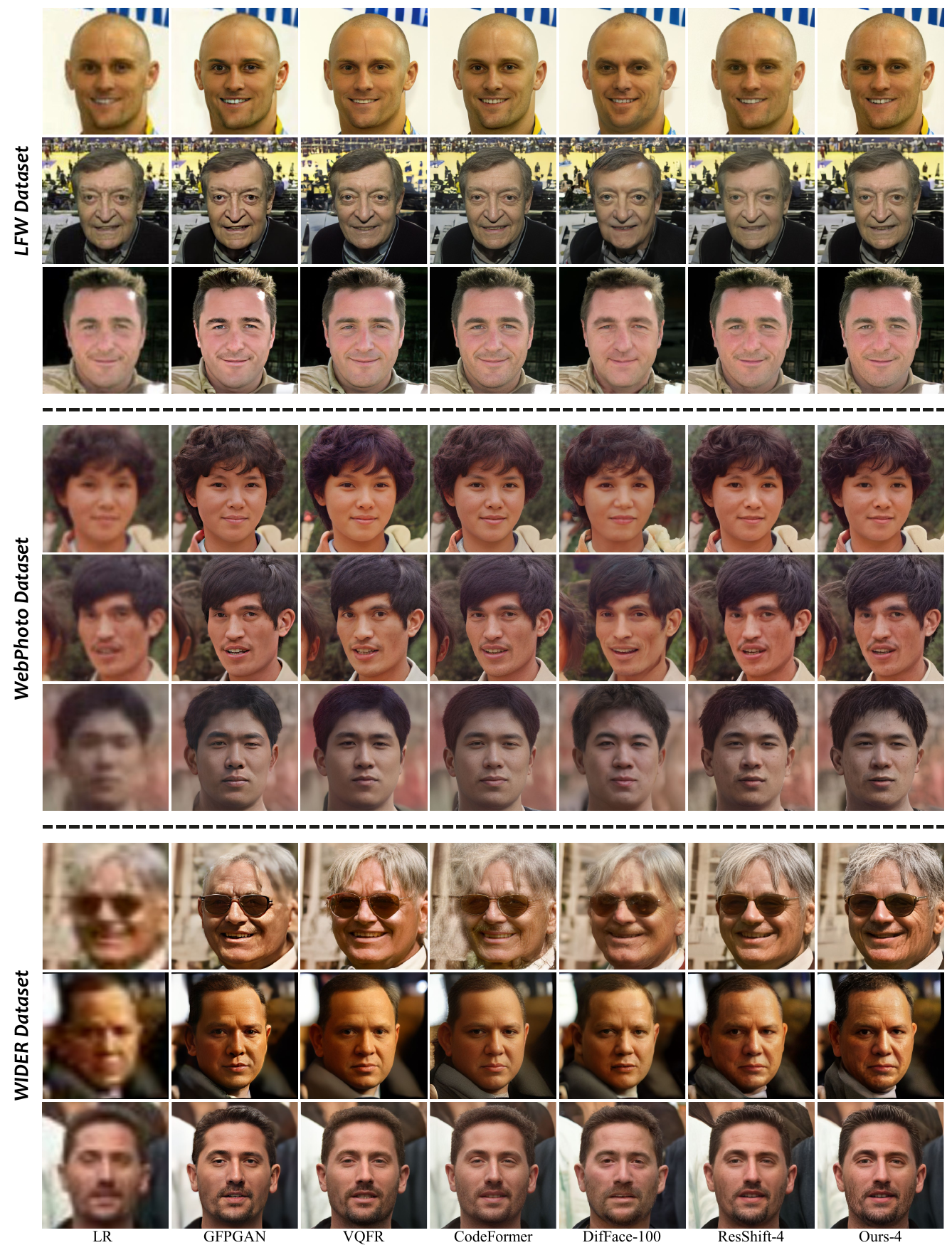}
	\caption{Qualitative results of different methods on three real-world datasets for blind face restoration. Please zoom in for a better view.
	}
	\label{fig:face_real_world}
\end{figure*}

\paragraph{Evaluation Metrics.}
To comprehensively assess various methods, this study adopts several reference-based metrics including LPIPS~\cite{zhang2018unreasonable}, identity score (IDS), landmark distance (LMD), and FID~\cite{heusel2017gans}. Non-reference metrics such as CLIPIQA~\cite{wang2023exploring}, MUSIQ~\cite{ke2021musiq}, and MANIQA~\cite{yang2022maniqa} are also employed.

\paragraph{Comparisons with State-of-the-Art Methods.}
Both synthetic dataset and real-world datasets are evaluated for blind face restoration. We present quantitative metrics in Table~\ref{tab:celeba_testing} Table~\ref{tab:real_face_testing}. Furthermore, we provide several real-world blind face restoration examples in Fig.~\ref{fig:face_real_world} and synthetic blind restoration face restoration examples in Fig.~\ref{fig:face_celeb}.

\end{document}